\documentclass[10pt, a4paper]{article}

\usepackage[final]{lrec2026} % this is the new style
\usepackage{latexsym}
\usepackage{booktabs}
\usepackage{multirow}
\usepackage{enumitem}
\usepackage{CJKutf8}
\usepackage[ruled,vlined]{algorithm2e}
\usepackage{makecell}
\usepackage{amsmath}

\title{ConceptKT: A Benchmark for Concept-Level Deficiency Prediction in Knowledge Tracing}

\name{Yu-Chen Kang, Yu-Chien Tang, An-Zi Yen} 

\address{
        Department of Computer Science, National Yang Ming Chiao Tung University, Taiwan \\
        \texttt{\{connie.cs12,tommytyc.cs10\}@nycu.edu.tw, azyen@nycu.edu.tw}
         }

\abstract{
Knowledge Tracing (KT) is a critical technique for modeling student knowledge to support personalized learning. However, most KT systems focus on binary correctness prediction and cannot diagnose the underlying conceptual misunderstandings that lead to errors. 
Such fine-grained diagnostic feedback is essential for designing targeted instruction and effective remediation.
In this work, we introduce the task of concept-level deficiency prediction, which extends traditional KT by identifying the specific concepts a student is likely to struggle with on future problems.
We present ConceptKT, a dataset annotated with labels that capture both the concepts required to solve each question and the missing concepts underlying incorrect responses.
We investigate in-context learning approaches to KT and evaluate the diagnostic capabilities of various Large Language Models (LLMs) and Large Reasoning Models (LRMs). 
Different strategies for selecting informative historical records are explored. % To mitigate the context length limitations of these models, we explore strategies for selecting informative historical records. 
Experimental results demonstrate that selecting response histories based on conceptual alignment and semantic similarity leads to improved performance on both correctness prediction and concept-level deficiency identification.
 \\ \newline  }

\begin{document}

\maketitleabstract

\section{Introduction}\label{sec:intro}

With the proliferation of online education platforms becoming mainstream learning environments, knowledge tracing (KT)~\citep{anderson1990cognitive} has emerged as a foundational technique for monitoring learners' evolving knowledge states and supporting personalized learning interventions. 
Specifically, KT seeks to predict a student's latent knowledge state by modeling their historical response records, thereby enabling the dynamic adaptation of instructional strategies and learning resources.
% Traditional KT approaches, such as Bayesian Knowledge Tracing (BKT)~\cite{corbett1994knowledge} and Deep Knowledge Tracing (DKT)~\cite{piech2015deep}, model student learning behaviors by processing sequential interaction data, typically encoded as tuples of question identifiers and binary correctness indicators.
% These models employ Hidden Markov Model (HMM)~\cite{baker2008more} or Long Short Term Memory (LSTM)~\cite{hochreiter1997long} to capture the dynamic evolution of a student's latent knowledge state over time.
Prior work~\cite{corbett1994knowledge,piech2015deep,pandey2019self,ghosh2020context,pandey2020rkt} in KT has primarily focused on leveraging deep sequential models and attention mechanisms to capture students' evolving knowledge states, with the goal of predicting their overall answer correctness.
Memory-augmented models~\citep{zhang2017dynamic,abdelrahman2019knowledge,liu2019ekt,wang2020neural} have been proposed to incorporate exercise content and concept-level information via memory-augmented architectures to infer a student-concept mastery matrix.
Other  approaches~\citep{zhang2024mlc,cui2023fine,luo2024efficient,qin2025interpretable} enhance interpretability through context-aware attention, relational modeling, hierarchical concept structuring, or difficulty-aware mechanisms.
However, most works focus on binary correctness prediction, with limited attention to the prediction of concept-level deficiencies.
% However, these methods still fall short of proactively identifying which specific concepts a student is most likely to misunderstand, particularly in items involving multiple concepts.

% While prior methods have demonstrated strong performance in predicting overall outcomes such as response correctness or performance categories, they generally lack the granularity required to anticipate concept-level deficiencies reflected in a student's problem-solving process.

\begin{table}[t]
    \centering
    \scriptsize
    \begin{tabular}{p{7.5cm}}
        \toprule
        \textbf{Question:} A welder received an order to make a 1 million liter cube - shaped tank . If he has only $4 \times 2$ meter sheets of metal that can be cut , how many metal sheets will be required for this order r ? ( 1 cubic meter = 1000 liters ) \\
        \textbf{Student Process:}
        \[
            1000000 L = 1000 m^3
        \]
        \[
            1000 \div 8 = 25
        \]
        \textbf{Correctness:} Wrong \\
        \midrule
        \textbf{Associated Concepts:} Volume Calculation, Area Calculation, Unit Conversion \\
        \textbf{Error Type:} Wrong Mathematical Operation/Concept \\
        \textbf{Error Equation:}
        \[
            1000 \div 8 = 25
        \]
        \textbf{Missing Concepts:} Volume Calculation, Area Calculation \\
        \textbf{Teacher Feedback:} The concept is incorrect. The metal sheets are only used for the six surfaces of the tank. Therefore, you should calculate the area of each surface and then calculate the number of metal sheets required, instead of using the total volume. The area of each surface is $1000 = 10^3$, $10 \times 10 = 100$. Therefore, the number of metal sheets required is $100 \div 8 \times 6 = 75$. \\
        \bottomrule
    \end{tabular}
    \caption{Example of a ConceptKT Instance Used for Answer Correctness and Concept-Level Deficiency Prediction based on Student Solution Processes.}\label{tab:ex_concept_diagnosis}
\end{table}

In practical educational scenarios, the utility of correctness prediction is enhanced when accompanied by insight into the specific concepts in which the student is likely to struggle.
Such fine-grained diagnostic feedback is essential for informing personalized instruction, designing targeted remedial materials, and implementing effective scaffolding strategies.
% Accordingly, advancing KT systems toward concept-level deficiency prediction represents a promising direction for improving the diagnostic capacity of intelligent tutoring systems.
Taking Table~\ref{tab:ex_concept_diagnosis} as an example, a student is required to first convert volume from liters to cubic meters, then compute the surface area of a cube, and finally estimate the number of metal sheets needed based on their area. 
The student's solution directly divides the volume in cubic meters by the area of a metal sheet, indicating a lack of understanding in both volume and surface area calculations, which ultimately leads to an incorrect answer.
Beyond predicting whether a student will answer correctly, the ability to anticipate specific concept-level deficiencies can significantly enhance the design of diagnostic assessments and adaptive instruction.
By anticipating concept-level deficiencies, instructional systems can adaptively select assessment items that align more closely with each student's learning needs. 
This strategy enhances the effectiveness of adaptive testing by targeting concepts where the student is likely to struggle, thereby improving diagnostic precision and instructional relevance.
For instance, if a system predicts that a student is likely to struggle with surface area calculation, it can provide preparatory exercises or modify the task flow accordingly. 
% Hence, to enhance the predictive granularity of traditional KT, we introduce an augmented task that not only predicts the correctness of a student's future response, but also identifies the specific conceptual deficiency that may contribute to an incorrect answer.

In this work, we construct ConceptKT by extending the MathEDU dataset~\cite{hsu2025mathedu} with concept-level annotations provided by three experts in mathematics education.
MathEDU consists of 4,048 solution process records from 6 students with diverse academic backgrounds, including Applied Mathematics, Finance, Japanese, Information Management, Mathematics Education, and Physics.
Each record includes rich error-related annotations, such as error type (e.g., incorrect operations or conceptual misunderstandings), error equation (the specific erroneous step), and teacher feedback that addresses the student's misconceptions. 
% To enhance the dataset's utility for concept-level prediction tasks, 
As shown in Table~\ref{tab:ex_concept_diagnosis}, we further label two types of annotations
(1) the \textbf{Associated Concepts} required to solve the problem, and
(2) the \textbf{Missing Concepts} that the student failed to demonstrate mastery of when answering incorrectly.
% An example of these annotations is shown in Table~\ref{tab:ex_concept_diagnosis}.
% These labels facilitate the development of models capable of identifying the specific conceptual deficiencies that may contribute to future errors.

Recently, large language models (LLMs) and large reasoning models (LRMs) have demonstrated remarkable capabilities in task understanding and reasoning.
An increasing number of studies show that LMs have strong potential in educational contexts, including math problem solving~\citep{yang2023large, didolkar2024metacognitive}, student feedback generation~\citep{hsu2025mathedu, baral2024automated}, and even knowledge tracing~\citep{cho2024systematic}.
Therefore, this study examines the capability of LMs to support concept-level diagnosis by extending KT beyond correctness prediction to include the identification of likely conceptual deficiencies through in-context learning.
Note that KT often requires models to process and integrate long sequences of student responses to accurately capture the evolution of their knowledge states.
Incorporating extensive historical data can introduce information redundancy or contextual noise, which may hinder the model's ability to effectively represent and reason about the student's current knowledge.
% In our setting, student problem-solving histories are provided as input to LLMs. 
% Each record is tokenized, and due to the context length limitations of LLMs, it is often infeasible to input the entire sequence at once.
This raises two research questions: 

\noindent \textbf{RQ1:} Should the entire history be fed into the model, or is it more effective to select only a subset of responses? 

\noindent \textbf{RQ2:} If selection is required, which records are most informative for predicting correctness and concept-level deficiencies?

To answer these questions, we explore strategies for selecting prior responses according to the concepts required by the target question, in order to predict both the correctness of a student's response and the specific concept deficiency underlying an incorrect answer.
% We also examine the effectiveness of large reasoning models (LRMs) in this task.
% We analyze their behaviors and summarize the practical limitations they exhibit.
In sum, our contributions are threefold:
% \begin{enumerate}[nolistsep]
    (1)~To support fine-grained instructional decisions in KT, we introduce an augmented task, i.e., concept-level deficiency prediction, that extends beyond correctness prediction by identifying the concept a student is likely to struggle with.
    (2)~We present a novel dataset, ConceptKT,\footnote{\url{https://github.com/NYCU-NLP-Lab/ConceptKT}} with expert-annotated concept-level labels that capture both the required concepts for each problem and the concepts in which students failed to apply correctly in their erroneous solutions.    
    (3)~We evaluate various LMs within an in-context learning paradigm for knowledge tracing, and analyze response history selection strategies. 
    The experiments demonstrate promising effects of selecting prior responses by conceptual relevance to the target question.
% \end{enumerate}

\section{Related Work}

\subsection{LLM-Enhanced and Open-Ended Knowledge Tracing}

Previous studies on knowledge tracing (KT) have primarily focused on deep sequential and attention-based architectures, such as DKT~\citep{piech2015deep}, SAKT~\citep{pandey2019self}, and AKT~\citep{ghosh2020context}, which model students’ response histories to capture temporal dependencies.
Memory-augmented approaches, including DKVMN~\citep{zhang2017dynamic} and EKT~\citep{liu2019ekt}, extend these models to track concept-level mastery with external memory components.
Graph-based extensions such as GKT~\citep{nakagawa2019graph}, AGKT~\citep{long2022automatical} and DyGKT~\citep{cheng2024dygkt} further incorporate structural relationships among concepts for enhanced reasoning.
While these models achieve strong correctness prediction, they provide limited diagnostic interpretability at the concept level.

% In recent years, 
% LLMs have achieved significant breakthroughs in natural language understanding and reasoning tasks.
Recently, LLMs have been applied to educational tasks involving knowledge tracing.
LLM-KT~\citep{wang2025llm} uses a pre-trained language model (e.g., BERT~\citep{devlin2019bert} or LLaMA~\citep{touvron2023llama}) to convert questions and concepts into embeddings, and employs traditional sequence learning models %(e.g., DKT, AKT) 
to encode question IDs and concept IDs.
These embeddings are then injected into the prompt of an LLM to capture both students' sequential behavioral patterns and the semantic features of the question text.
LLMKT~\citep{scarlatos2025exploring} is applied to tutor-student dialogue scenarios. 
It utilizes prompt-based techniques to identify the knowledge components involved in student responses and provides real-time assessments of their mastery levels.
DDKT~\citep{cen2025llm} enables LLMs to perform step-by-step reasoning over questions and generate solution processes in order to estimate difficulty, it also calculates statistical difficulty based on students' historical correctness rates.
By combining both the difficulty estimated by the LLMs and the statistical difficulty derived from performance data, DDKT~\citep{cen2025llm} enhances the predictive accuracy in KT tasks.
KnowTrace~\citep{li2025knowtrace} and SINKT~\citep{fu2024sinkt} focus on knowledge structure and graph-based modeling.
In addition, several studies have begun to explore the application of knowledge tracing to open-ended problems, such as OKT~\citep{liu2022open} and ECKT~\citep{yu2024eckt}.

\subsection{Mathematics Education Datasets}

Several datasets used in mathematics education research have been constructed.
The ASSISTments datasets\footnote{\url{https://www.etrialstestbed.org/resources/featured-studies/dataset-papers}} are derived from the online mathematics tutoring platform.
It includes multiple versions of student interaction records, covering secondary school mathematics curricula in the United States.
% Due to its large scale and high-quality annotations, the ASSISTments series has become one of the most widely used benchmark datasets in KT research.
The Junyi dataset~\cite{pojen2020junyi} comes from the learning platform Junyi Academy. %in Taiwan.
The data was collected after 2016 and covers mathematical concepts and problem types from elementary to junior high school.
Eedi2020~\cite{wang2020instructions} is a multiple-choice dataset released by the educational platform Eedi. %in the UK.
The questions are designed for students in Grades 7-9.
The Algebra datasets, presented at the KDD Cup 2010 Educational Data Mining Challenge~\cite{stamper2010challenge}, target secondary school students (Grades 8–10) studying algebra.
% The Algebra datasets, presented at the KDD Cup 2010 Educational Data Mining Challenge~\cite{stamper2010challenge}, originate from the Carnegie Learning Algebra system in the United States.
% Algebra 2005-06 was collected from 2005 and 2006, Algebra 2006-07 and Bridge to Algebra were collected from 2006 to 2007, targeting secondary school students (Grades 8-10) studying algebra.
% EdNet~\cite{choi2020ednet} is released in 2020 by the developer of the QANDA platform.
EdNet~\cite{choi2020ednet} contains student interaction data from 2018 to 2020, mainly focusing on secondary school mathematics.
% DBE-KT22~\cite{abdelrahman2022dbe} is a dataset released during the 2022 knowledge tracing competition organized by DBE (Data-driven Basic Education) in China, it 
DBE-KT22~\cite{abdelrahman2022dbe} covers responses from junior high school students.
% These datasets all include student responses to each problem, associated knowledge components, timestamps, and correctness.

DrawEduMath~\cite{baral2025drawedumath} is collected around 2022 from the DrawEdu platform. %in China.
It contains data from students in Grades 7-12 solving interactive geometry drawing problems.
\citet{erickson2020automated} constructed a dataset comprising student responses to open-ended mathematics questions, sourced from open educational resources such as EngageNY, Illustrative Mathematics, and Utah Math.
The questions were manually graded by teachers, and each student response is labeled as ``Correct,'' ``Partially Correct,'' or ``Incorrect.''
% While both datasets include students' solving processes, they do not provide associated knowledge component annotations.
Previous datasets used for KT have included annotations of the corresponding knowledge concepts for each question.
However, most of these datasets only record the final responses of students, lacking detailed traces of the reasoning steps involved in their problem-solving processes.
MathEDU is a dataset derived from the MathQA dataset~\cite{amini2019mathqa}, focused on word-based math problems at the secondary school level.
Students from diverse backgrounds were invited to write detailed reasoning and solutions.
The dataset includes students' step-by-step responses, correctness annotations, and error type labels, such as conceptual misunderstandings and calculation errors. %, making it ideal for research on cognitive error modeling.
% Such information is highly valuable for LLMs, which possess strong language understanding capabilities, as it enables them to better capture students' knowledge states.
Thus, we extend the MathEDU dataset into ConceptKT by incorporating concept-level annotations. %, enabling more deeper analysis of students' cognitive states.

\begin{table}[t]
  \centering
  \scriptsize
  \setlength\tabcolsep{1mm}
  \begin{tabular}{lcccc}
    \toprule
          &       & \multicolumn{1}{c}{\multirow{3}[0]{*}{\shortstack{\textbf{Correct} \\ \textbf{Answers}}}} & \multicolumn{2}{c}{\textbf{Wrong Answers}} \\
          & \textbf{\#Ans.} &       & \multicolumn{1}{c}{\textbf{Careless}} & \multicolumn{1}{c}{\textbf{Concept-Level}} \\
          &       &       & \multicolumn{1}{c}{\textbf{Mistake}} & \multicolumn{1}{c}{\textbf{ Deficiencies}} \\
    \midrule
    S1 & 683 & 70.57\% & 18.01\% & 11.42\% \\
    S2 & 685 & 87.59\% & 7.45\% & 4.96\% \\
    S3 & 678 & 71.09\% & 13.28\% & 15.63\% \\
    S4 & 660 & 75.30\% & 15.00\% & 9.70\% \\
    S5 & 682 & 67.01\% & 17.30\% & 15.69\% \\
    S6 & 660 & 80.61\% & 8.18\% & 11.21\% \\
    \midrule
    Overall & 4,048 & 75.35\% & 13.21\% & 11.44\% \\
    \bottomrule
  \end{tabular}
  \caption{Distribution of Student Responses.}
  \label{tab:correctness_of_students}
\end{table}

\section{Dataset Construction and Analysis}
\subsection{From MathEDU to ConceptKT}\label{sec:conceptkt}
% MathEDU was originally developed to evaluate the capability of LLMs in assessing student responses.
MathEDU includes annotated error types related to students' problem-solving processes:

\noindent \textbf{Wrong mathematical operation/concept:} Student applies an incorrect mathematical operation or uses an inappropriate mathematical concept to solve a problem.  

\noindent \textbf{Lack of necessary mathematical concepts:} Errors in answering caused by a lack of essential mathematical knowledge or techniques.  

\noindent \textbf{Calculation Error:} Mistakes in calculations, such as errors in solving equations, arithmetic mistakes, and incorrect unit conversions.  

\noindent \textbf{Incomplete Answer:} Student used a correct formula or procedure but did not complete it.  

\noindent \textbf{Careless Error:} Errors caused by students' carelessness in answering, including number substitution errors and missing digits.  

We consider ``Wrong Mathematical Operation/Concept'' and ``Lack of Necessary Mathematical Concepts'' as indicators of \textbf{conceptual deficiency}, while the remaining three error types are regarded as \textbf{careless mistakes}.
Thus, we focus on the two concept-related error types and annotate the corresponding data with concept labels.

% \subsubsection{Student Problem-Solving Performance}
To characterize students' problem-solving performance, Table~\ref{tab:correctness_of_students} reports the proportions of correct and incorrect problem-solving results for each individual.
% Among the incorrect responses, errors are further categorized into careless mistakes and concept-level deficiencies as mentioned in Section~\ref{sec:conceptkt}.
% Students 1, 3, and 5 exhibit relatively higher overall error rates, with Students 3 and 5 particularly prone to concept-level deficiencies.
Students exhibit different error patterns. While most make more careless mistakes, some (e.g., Students 3 and 6) show greater concept-level deficiencies, indicating deeper conceptual gaps.
These findings highlight that modeling students' learning states requires not only predicting correctness but also distinguishing error types to reveal their underlying causes.
% For most students, careless mistakes occur more frequently than concept-level deficiencies. 
% However, Students 3 and 6 show the opposite pattern, where concept-level deficiencies exceed careless mistakes, suggesting that their difficulties are more likely due to gaps in conceptual understanding rather than superficial lapses.
% These findings highlight the challenge of the proposed task: to effectively model students' learning states, it is not sufficient to predict correctness alone. 
% It is also essential to distinguish between types of errors in order to uncover their underlying causes.

\subsection{Data Annotation}

Based on the U.S. Common Core State Standards for mathematics\footnote{\url{https://corestandards.org/mathematics-standards/}} for K-12 students and the questions in MathEDU, we define 55 core mathematical concepts.
% Each concept is accompanied by detailed annotation guidelines, as shown in Appendix~\ref{sec:concept}.
Three experts were invited in mathematics education to review each student's problem-solving process and annotate both the associated concepts assessed by the problem and the missing concepts demonstrated by the student's errors. %Each expert annotator was compensated at a rate above the minimum wage in their respective country.
Before annotation, they were instructed to carefully review the guideline.
% For questions labeled in MathEDU as ``Wrong Mathematical Operation/Concept'' or ``Lack of Necessary Mathematical Concepts'', the annotators were instructed to examine the student's problem-solving process and identify the missing concepts that contributed to the incorrect answer.
Both associated concepts and missing concepts were annotated using a multi-label format, as each problem may involve multiple relevant concepts and a student's error may arise from multiple conceptual deficiencies.

To assess annotation quality, we computed Fleiss' $\kappa$ to evaluate inter-annotator agreement among the three experts.
The $\kappa$ score was 0.6799 for associated concepts and 0.6328 for missing concepts, indicating substantial agreement according to standard interpretation thresholds.
In cases of annotation disagreement, we applied a majority voting scheme.
Each label was determined by the majority of annotators.
When majority voting failed to resolve uncertainty, the annotators engaged in collaborative discussion until consensus was reached.
As a result, a total of 4,048 records were annotated. Each question was labeled with an average of 1.2441 associated concepts. 
Among these, 463 records involved incorrect responses requiring the annotation of missing concepts, with an average of 1.112 missing concepts per question.

\subsection{Dataset Statistics and Analysis}

\begin{figure}
    \centering
    \includegraphics[width=\linewidth]{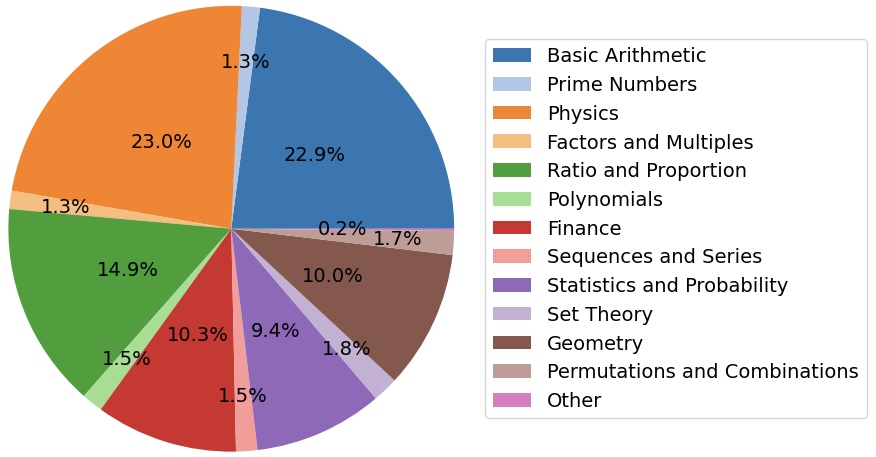}
    \caption{Distribution of Questions Across 13 Categories.}
    \label{fig:topic_statistic}
\end{figure}

\begin{figure*}
    \centering
    \includegraphics[width=\linewidth]{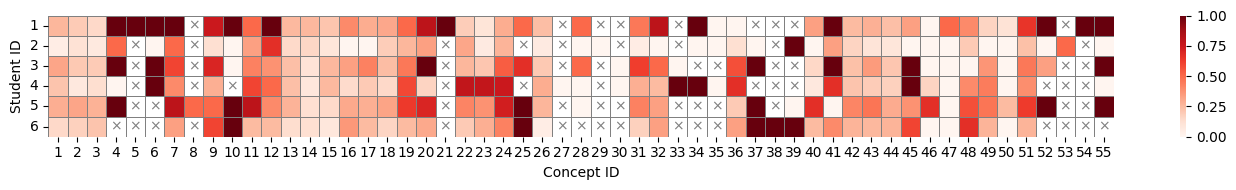}
    \caption{Concept-level error rates across students, where darker shades indicate higher error proportions and ``×'' marks denote concepts that were not covered in the student's problem-solving history. 
    %Concept ID mappings are listed in Appendix~\ref{sec:concept}.
    }
    \label{fig:error_distribution}
\end{figure*}

\subsubsection{Concept Statistics}\label{sec:concept_stat}
In consultation with the experts, we further grouped the 55 defined concepts into 13 high-level categories, including ``Basic Arithmetic,'' ``Prime Numbers,'' ``Factors and Multiples,'' ``Physics,'' ``Ratio and Proportion,'' ``Finance,'' ``Statistics and Probability,'' ``Polynomials,'' ``Sequences and Series,'' ``Geometry,'' ``Set Theory,'' ``Permutations and Combinations,'' and ``Other''. % along with an additional ``Other'' category, resulting in a total of 13 categories.
% The full mapping between categories and their associated concepts is provided in Appendix~\ref{sec:concept}.
The 13 categories contain the following 55 concepts, each associated with its respective ID:

\noindent \textbf{Basic Arithmetic:} 1. Basic Math Operations, 2. Expression and Operations with Symbols, 3. Unit Conversion, 4. Range of Values, 5. Modular Arithmetic, 6. Factorial

\noindent \textbf{Prime Numbers:} 7. Prime, 8. Composite Number, 9. Prime Factorization

\noindent \textbf{Factors and Multiples:} 10. Number of Factors, 11. Greatest Common Divisor, 12. Least Common Multiple

\noindent \textbf{Physics:} 13. Distance-Time-Speed, 14. Relative Speed, 15. Workload-Time-Speed, 16. Mixed Solution Concentration

\noindent \textbf{Ratio and Proportion:} 17. Direct Proportion and Inverse Proportion, 18. Ratio Calculation

\noindent \textbf{Finance:} 19. Simple Interest, 20. Compound Interest, 21. Effective Annual Interest Rate, 22. Profit, 23. Loss, 24. Discount Problem, 25. Price Calculation

\noindent \textbf{Statistics and Probability:} 26. Arithmetic Mean, 27. Mode, 28. Median, 29. Standard Deviation, 30. Normal Distribution, 31. Probability

\noindent \textbf{Polynomials:} 32. Square of the Sum, 33. Difference of Squares, 34. Sum of Squares, 35. Factorization of Polynomials, 36. Function, 37. Roots and Coefficients, 38. Quadratic Equation, 39. Absolute Value Equation

\noindent \textbf{Sequences and Series:} 40. Arithmetic Sequence/Series, 41. Geometric Sequence/Series

\noindent \textbf{Geometry:} 42. Perimeter Calculation, 43. Area Calculation, 44. Volume Calculation, 45. Graphs of Cartesian Coordinates and Linear Equations, 46. Pythagorean Theorem, 47. Solid Geometry, 48. Plane Geometry

\noindent \textbf{Set Theory:} 49. Inclusion-Exclusion Principle, 50. Fundamental Counting Principles

\noindent \textbf{Permutations and Combinations:} 51. Permutations and Combinations, 52. Pigeonhole Principle

\noindent \textbf{Other:} 53. Exact Value, 54. Local Value, 55. Face Value

Figure~\ref{fig:topic_statistic} shows the distribution of questions across these categories.
Among them, ``Basic Arithmetic'', ``Physics'', and ``Ratio and Proportion'' constitute the largest proportions, suggesting that these categories encompass a broad range of problem types and are frequently encountered in ConceptKT.
% For example, ``Physics'' category includes applications of formulas such as $Distance = Speed \times Time$, calculations involving relative motion, and problems concerning solution concentration.
In contrast, categories such as ``Prime Numbers'' and ``Factors and Multiples'' each account for less than 2\% of the dataset.

\subsubsection{Student Concept Mastery Analysis}

To better understand each student's concept mastery, we present a heatmap in Figure~\ref{fig:error_distribution}. where the vertical axis corresponds to the six student IDs and the horizontal axis indicates the concept IDs.
% A list of concept-ID mappings is provided in Appendix~\ref{sec:concept}.
Each cell in the heatmap represents the error rate of a student on a specific concept.
Darker colors indicate higher error rates, corresponding to lower levels of conceptual understanding.
Cells marked with ``×'' indicate that the student did not encounter any problems involving the corresponding concept.

% As shown in Figure~\ref{fig:error_distribution}, 
Students 1, 3, and 5 exhibit high error rates in Concept IDs 1–6 (i.e., ``Basic Arithmetic''), suggesting a weaker mathematical foundation that may have negatively impacted their performance across other concepts as well.
Their errors span a wide range of concepts, indicating generally lower mastery across multiple areas. % Their errors are distributed across a broad range of concepts, indicating not only higher overall error rates but also generally lower mastery across multiple conceptual areas.
In contrast, Student 2 demonstrates relatively strong performance across most concepts.
Student 4 shows lower performance in Concept IDs 19–25 (i.e., ``Finance'') and Concept IDs 32–39 (i.e., ``Polynomials'').
Student 6 also struggle with Polynomials.

\begin{figure}
    \centering
    \includegraphics[width=\linewidth]{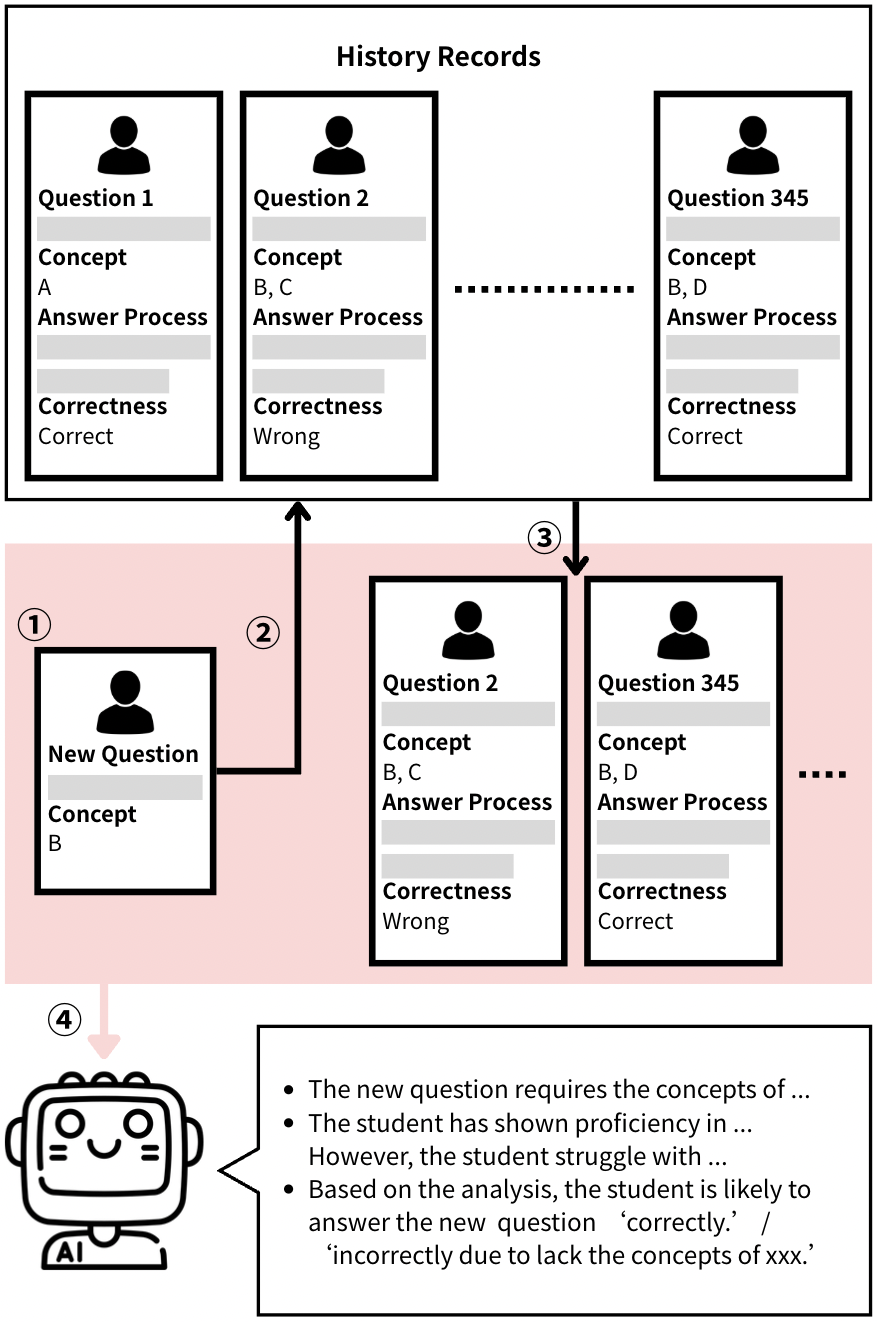}
    \caption{Overview of Knowledge Tracing.}
    \label{fig:kt}
\end{figure}

\section{Methodology}
\subsection{Task Formulation}

% \begin{figure}
%     \centering
%     \includegraphics[width=\linewidth]{figures/Framework.png}
%     \caption{Overview of Knowledge Tracing.}
%     \label{fig:framework}
% \end{figure}

This study predicts a student's performance on new math questions and identifies potential concept deficiencies based on their problem-solving history. % This study focuses on predicting a student's performance on a new mathematical question based on their prior problem-solving history, with an additional objective of identifying potential deficiencies in relevant mathematical concepts.
We adopt an in-context learning framework with LLMs to perform this extended KT task.
To guide the model's reasoning process, we incorporate a prompt-based Chain-of-Thought (CoT) strategy, in which structured instructions are provided to elicit step-by-step conceptual analysis.
% This involves (1)~identifying the key mathematical concepts required by the target question, and (2)~analyzing the student's previous responses to assess their conceptual mastery.
Figure~\ref{fig:kt} shows the overview of our task.

Specifically, let $q_i$ denote the $i$-th question, $r_i$ denote the student's solution process for $q_i$, and $c_i \in$ \{\textit{correct}, \textit{wrong}\} indicate the corresponding correctness label.
For the target question $q$, the student's response history $H$ = \{($q_1$, $r_1$, $c_1$),\dots, ($q_n$, $r_n$, $c_n$)\} is presented to the model $\mathcal{M}$ as few-shot examples.
The model is instructed to perform three steps:
(1)~\textbf{Associated Concept Identification:} infer the associated concepts required to solve the target question $q$, with the resulting analysis denoted as $\alpha$;
(2)~\textbf{Student Modeling:} analyze the student's response history $H$ to assess their concept-level mastery, with the resulting analysis denoted as $\beta$;
(3)~\textbf{Answer Correctness and Concept-level Deficiency Prediction:} predict the student's response correctness $c$ for the target question $q$, and identify the set of concept-level deficiencies $\gamma$ that may contribute to an incorrect response.
The complete process can be expressed as $(\alpha, \beta, c, \gamma) = \mathcal{M}(H,q)$.

% In KT tasks, students continuously accumulate interaction records as they progress through their learning.
% However, LLMs are typically constrained by limited context length, making it impractical to input a student's entire interaction history.
% To address this constraint, it becomes necessary to selectively sample a subset of prior records that are most informative for predicting the student's future performance.
However, not all prior problem-solving records are equally relevant to the target question. Therefore, it is necessary to selectively sample the most informative ones for predicting a student's future performance.
Formally, given a response history $H$ consisting of 
$n$ interactions, the objective is to select a subset of $m$ records ($m < n$) to support effective correctness and Concept-level Deficiency Prediction.
% To address the two research questions outlined in Section~\ref{sec:intro}, we investigate the impact of historical response selection under two distinct scenarios.
In our experiments, models capable of encoding up to 110K tokens are sufficient to fully accommodate a student's entire response history.
We use models with sufficient context length (up to 1 million tokens), allowing the entire student history to be included. 
We investigate whether selective record inclusion remains beneficial.

% To explore how to best support prediction when not all responses can be included, we explore four different selection strategies. 
% These strategies are evaluated based on their effectiveness in improving both answer correctness prediction and concept-level deficiency identification.

\subsection{Response Selection Strategies}\label{sec:selection_strategies} %Strategies for Response History Selection

% We consider the following principles to design selection strategies:

% \noindent \textbf{Random Sampling:}
% A simple baseline in which $m$ responses are randomly sampled from the student's all $n$ problem-solving results.

% \noindent \textbf{Concept Relevance:}
% Assuming that examples involving similar concepts offer more relevant context for reasoning about the student's conceptual mastery, we select the response that covers one or more of the same concepts required by the target question $q$.
The main principle behind the selection strategies is that examples involving similar concepts provide more relevant context for reasoning about a student's conceptual mastery. 
Accordingly, we select prior responses that cover one or more of the same concepts required by the target question $q$.
The following selection strategies are tested:

\noindent \textbf{(1)~All Responses:}
All responses of the student are used without any filtering or selection.

\noindent \textbf{(2)~Same-Concept Only:}
All responses that involve the same concepts as $q$ are selected.

\noindent \textbf{(3)~Conceptual Semantic Selection:}  
This strategy prioritizes prior responses that are both conceptually and semantically relevant to $q$.  
We first identify the subset of records in the student's history $H$ that involve one or more of the same concepts as $q$, yielding $k$ candidate responses.  
If $k \leq m$, all available responses are used.  
If $k > m$, we compute the semantic similarity between each candidate question and $q$ using BERTScore~\citep{zhang2019bertscore}, and select the top $m$ responses with the highest similarity.  
We limit the number of responses to $m$.
If fewer than $m$ responses are available, only those records are used. % When fewer than $m$ responses are available (i.e., $k < m$), only the available $k$ records are used without additional supplementation.

\begin{table}[t]
  \centering
  \footnotesize
    \begin{tabular}{lccc}
    \toprule
          & Total & Training Set & Test Set \\
    \midrule
    Student 1 & 683   & 615   & 68 \\
    Student 2 & 685   & 616   & 69 \\
    Student 3 & 678   & 610   & 68 \\
    Student 4 & 660   & 594   & 66 \\
    Student 5 & 682   & 614   & 68 \\
    Student 6 & 660   & 594   & 66 \\
    \midrule
    Total & 4,048   & 3,643   & 405 \\
    \bottomrule
    \end{tabular}%
  \caption{Number of Questions in the Training and Test Sets.}
  \label{tab:data_splitting}%
\end{table}%

\section{Experiments}
\subsection{Experimental Setup}

We experiment with various LLMs and LRMs on our task, with parameter sizes ranging from 7B to 671B.
All models are configured with a fixed temperature of 0 to ensure reproducibility.
A unified prompt template is used for the KT task across all models.
Adaptive prompt tuning tailored to each model is left for future work.

Table~\ref{tab:data_splitting} shows the number of questions in the training and test sets. 
The training set is used for selecting historical responses to represent student knowledge mastery, while the test set provides the target questions used for evaluation.
Table~\ref{tab:prompt} presents the prompt used to instruct the model to perform the KT task.

\begin{table}[t]
    \centering
    \small
    \begin{tabular}{p{7cm}}
        \toprule
        You will be provided with several sets of historical records, ending with a new question.
        
        \# Task Instructions:
        1. Identify the concepts required to answer the new question (Do not solve the new question).
        2. Compare these concepts with the student's past performance. \\
        3. Determine whether the student can answer the new question correctly. \\
            - If the student is likely to answer correctly, output: \\
            Correct. No lacking concepts. \\
            - If the student is likely to answer incorrectly, output the missing concepts using this strict format: \\
            Wrong. Lack of concept1 \&\& concept2 \\
            - Replace concept1, concept2, etc., with actual missing concepts from the **Concept List**. \\
            - If multiple concepts are missing, separate them using `` \&\& '' (no extra spaces, punctuation, or explanations). \\
            - Never use placeholders like concept1 \&\& concept2—only real concepts from the list. \\
        
        \# Concept List (Strict Selection Only): \\
        \{Concept list as specified in Section~\ref{sec:concept_stat}\}
        
        \# Historical Records:\\
        Question 1: \{Question\}\\
        Student Response: \{Response\}\\
        Student Response Correctness: \{Yes or No\}\\
        \\
        Question 2: \{Question\}\\
        Student Response: \{Response\}\\
        Student Response Correctness: \{Yes or No\}\\
        \\
        ...\\
        \\
        \# New Question: \{Target Question\}\\

        \bottomrule
    \end{tabular}
    \caption{KT Inference Prompt.}
    \label{tab:prompt}
\end{table}

\begin{table*}[t]
  \centering
  \resizebox{\linewidth}{!}{
    \begin{tabular}{llrccc}
    \toprule
          \textbf{Selection Strategy} & \textbf{Model} & 
          \multicolumn{1}{c}{\textbf{Correctness}} & 
          \multicolumn{3}{c}{\textbf{Missing Concept}} \\
          & & \textbf{Accuracy} & \textbf{Macro-F1} & \textbf{Macro-Precision} & \textbf{Macro-Recall} \\
    \midrule
    N/A & DKT   & 64.80\% & -      & -   & -    \\
    N/A & DKVMN & 63.10\% & -      & -   & -    \\
    N/A & GKT   & 63.20\% & -      & -   & -    \\
    N/A & SAKT  & 66.46\% & -      & -   & -    \\    
    N/A & OKT   & 69.25\% & 1.87\% & 2.07\% & 1.92\% \\
    \midrule
    All Responses & Gemini-2.0-Flash & 48.01\% & 14.08\% & 12.12\% & \underline{29.38\%}\\
     & Llama-4 & 68.81\% & 5.43\% & 5.66\% & 5.44\% \\
     & o3-mini & 70.35\% & 2.16\% & 2.67\% & 2.11\% \\
     & DeepSeek-R1 & 67.70\% & 14.82\% & 13.85\% & 19.06\% \\
    \midrule
    Same-Concept Only & Gemini-2.0-Flash & \textbf{60.83\%}& \textbf{15.99\%} & \textbf{13.63\%} & \textbf{29.66\%} \\
     & Llama-4 & \underline{73.67\%}& \underline{10.13\%}& \underline{12.07\%}& \underline{9.86\%}\\
     & o3-mini & \underline{70.35\%}& \underline{10.90\%}& \underline{11.36\%}& \underline{11.81\%}\\
     & DeepSeek-R1 & \textbf{71.02\%} & \textbf{17.40\%} & \underline{17.00\%}& \textbf{23.46\%} \\
    \midrule
    Conceptual Semantic Selection & Gemini-2.0-Flash & \underline{60.62\%}& \underline{15.55\%}& \underline{13.23\%}& 26.57\% \\
     & Llama-4 & \textbf{73.89\%} & \textbf{13.83\%} & \textbf{14.46\%} & \textbf{14.48\%} \\
     & o3-mini & \textbf{71.02\%} & \textbf{12.77\%} & \textbf{14.39\%} & \textbf{15.61\%} \\
     & DeepSeek-R1 & \underline{70.12\%}& \underline{16.87\%}& \textbf{18.63\%} & \underline{20.56\%}\\
    \bottomrule
    \end{tabular}%
  }
  \caption{Results of Answer Correctness and Concept-Level Deficiency Prediction.}
  \label{tab:experimental_results}%
\end{table*}%

% We sort each student's problem-solving records in the order they were answered, using the first 90\% as $H$ and the remaining 10\% as the test set.
% Specifically, we select records from the first 90\% as the student's historical input to the model, which is then used to predict their performance on the final 10\% of questions.
% For a student with $N$ total problem-solving records, we consider the first $n$ responses as $H$, which is provided to $\mathcal{M}$ as few-shot examples.
We chronologically order each student's problem-solving records and use the first 90\% as the response history $H$ and the remaining 10\% as the test set.
Formally, for a student with $N$ total responses, we define the first $n = \lfloor 0.9N+0.5 \rfloor$ records as $H = {(q_1, r_1, c_1), \dots, (q_n, r_n, c_n)}$, which are provided to the model $\mathcal{M}$ as few-shot examples.
The model is then tasked with predicting the answer correctness $c_{n+j}$ and concept-level deficiencies $\gamma_{n+j}$ for each target question $q_{n+j}$, where $j = 1, 2, \dots, N - n$.

% Since models with a context length of up to 1 million tokens can accommodate the full 90\% response history used as input, we categorize models into two experimental settings based on their context length capacity.
% For the selection-required setting, Qwen2.5-7B-Instruct~\citep{qwen2025qwen25technicalreport}, Mistral-Small-24B-Instruct,\footnote{\url{https://mistral.ai/news/mistral-small-3-1}} Llama-3.3-70B~\citep{grattafiori2024llama}, and QwQ-32B~\citep{yang2025qwen3} are used.
% Given the context length constraints of these models, we empirically determine that approximately 30 responses can be encoded before reaching the input limit.
% Accordingly, 
We set $m$ to 30 in our experiments, and further investigate the effect of varying $m$ values in the subsequent analysis.
Gemini-2.0-Flash,\footnote{\url{https://deepmind.google/technologies/gemini/flash-thinking/}} Llama-4,\footnote{\url{https://huggingface.co/meta-llama/Llama-4-Maverick-17B-128E-Instruct}} o3-mini,\footnote{\url{https://openai.com/index/openai-o3-mini/}} DeepSeek-R1~\cite{guo2025deepseek} are employed.
Additionally, we train DKT, DKVMN, GKT, SAKT, and OKT as baseline models, among which only OKT can predict concept deficiencies.

\subsection{Experimental Results}

We evaluate two tasks in this study: answer correctness prediction and concept-level deficiency prediction.
Accuracy is used as the evaluation metric for the answer correctness prediction.
Macro-F1 is adopted for the concept-level deficiency prediction, as it is a multi-label classification problem.

Table~\ref{tab:experimental_results} presents the performance of various LLMs and LRMs under different historical response selection strategies.
The best-performing result for each model across all selection strategies is highlighted in bold, while the second-best is underlined. 
Comparisons are made within each individual model (i.e., within-row grouping) to isolate the impact of different selection strategies.

OKT demonstrates strong accuracy in predicting answer correctness. 
However, it performs poorly in predicting concept-level deficiencies.
Comparing different strategies, using all responses without filtering (\textit{All Responses}) leads to a significant drop in concept-level deficiency prediction performance. % when all responses are used without filtering (All Responses), model performance on concept-level deficiency prediction degrades significantly. 
For example, Llama-4 and o3-mini achieve Macro-F1 scores of only 5.43\% and 2.16\%, respectively. 
In contrast, restricting the input to responses involving the same concepts as the target question (Same-Concept Selection) leads to substantial performance gains.
DeepSeek-R1 achieves the highest Macro-F1 score (17.40\%) for concept-level deficiency prediction among all models in this setting, along with strong performance on answer correctness prediction (71.02\%). 
These results indicate that even when a model is capable of ingesting the full student history, concept-aware input filtering remains critical for guiding model reasoning and maintaining diagnostic precision.

The results of ``Conceptual Semantic Selection'' show that this strategy further improves performance for certain models, confirming its effectiveness in selecting informative responses. % The results of ``Conceptual Semantic Selection''  demonstrate that this strategy yields additional performance improvements for certain models, confirming its effectiveness in selecting informative responses.
Using this strategy, o3-mini exhibits a remarkable improvement in concept-level deficiency prediction, rising from 2.16\% to 12.77\%, with a statistically significant difference ($t$-test, $p < 0.01$).
Hence, for models with reasoning capacity, carefully selecting a small set of conceptually relevant examples can be more beneficial than simply increasing the quantity of input history.

In summary, these findings suggest that simply increasing the number of input responses does not necessarily enhance model performance. 
Instead, the quality and conceptual relevance of the selected responses are key determinants of success in both answer correctness prediction and concept-level deficiency identification. 
Selection strategies that jointly consider conceptual alignment and semantic similarity enable models to more effectively reason about students' conceptual understanding.
Appropriate selection strategies enable small-scale LLMs to achieve promising performance even when historical responses are limited, as often seen in real-world settings.

We conducted a manual analysis of the model's predictive behavior and identified the primary source of difficulty.
A key challenge is the variability of student performance.
Student performance on target questions often diverges markedly from their prior responses.
This unpredictability is especially pronounced among students with moderate mastery levels, whose learning trajectories tend to be unstable.
Furthermore, a high frequency of non-conceptual errors (e.g., careless mistakes) in the historical responses can mislead the model into falsely inferring underlying conceptual deficiencies.
% The further analysis is presented in Appendix~\ref{sec:human_evaluation}.

\section{Discussion}\label{sec:discussion_2}
% \subsection{Impact of Response History Augmentation}\label{sec:discussion_2}

Due to the limited availability of response records for some concepts (as shown in Figure~\ref{fig:topic_statistic}), the conceptual semantic selection strategy may sometimes yield an insufficient number of historical responses for a given target question.
This lack of same-concept context can hinder the model's ability to accurately assess the student's conceptual understanding.
To investigate this challenge, we examine whether supplementing the input with additional responses involving different concepts can mitigate the sparsity of same-concept records and assist in maintaining prediction quality. 
Alternatively, such supplementation may introduce semantic noise and negatively affect model performance.

We restrict this analysis to cases where the number of same concept responses is fewer than 30, in order to assess the impact of supplementing the input with different-concept responses under data-sparse conditions.
A total of 106 target questions are included in this experiment.
The following three augmentation methods are evaluated:

\noindent \textbf{Conceptual Semantic Selection (No Augmentation):}
This setting follows the method described in Section~\ref{sec:selection_strategies}, % where the model receives all available responses involving the same concepts as the target question, 
without any additional augmentation from other concepts.

\noindent \textbf{Random Augmentation:}
After applying the conceptual semantic selection strategy, additional responses are appended by randomly sampling from the remaining responses involving different concepts, until the total reaches 30.

\noindent \textbf{Similarity-Based Augmentation:}
Similar to Random Augmentation, but instead of random sampling, responses from different concepts are selected based on their semantic similarity to the target question, measured using BERTScore.

We select Llama-4, which achieves the best performance under the conceptual semantic selection strategy in both selection-required and selection-optional settings.
As shown in Table~\ref{tab:filling_with_others}, Llama-4 performs better when using only same-concept responses (i.e., no augmentation), even if the total number of responses is less than 30. 
Supplementing with responses from different concepts, whether randomly or based on semantic similarity, tends to introduce contextual inconsistency, thereby reducing both correctness prediction accuracy and concept-level deficiency identification. 
These findings suggest that conceptual alignment is more critical than input quantity in supporting model reasoning and prediction reliability.

\begin{table}[t]
  \centering
  \small
  \begin{tabular}{lrr}
    \toprule
    \textbf{Augmentation} & \textbf{Correctness} & \textbf{Missing Concept} \\
    \midrule
    No                & \textbf{68.87\%} & \textbf{9.15\%} \\
    Random            & 66.98\%          & 5.08\% \\
    Similarity-Based  & 66.04\%          & 7.66\% \\
    \bottomrule
  \end{tabular}
  \caption{Results of Llama-4 under Different History Augmentation Strategies.}
  \label{tab:filling_with_others}
\end{table}

\section{Conclusion}
Knowledge tracing plays a crucial role in enabling personalized learning and adaptive instruction. 
However, existing research focuses on answer correctness prediction, with limited support for diagnosing concept-level deficiencies. 
Moreover, most datasets used in Mathematics KT studies lack students' authentic problem-solving processes. 
To address these limitations, we introduce a new task formulation that jointly predicts answer correctness and concept-level deficiencies, and systematically evaluate the capabilities of various LLMs and LRMs under an in-context learning setting.
We construct ConceptKT, the first expert-annotated dataset with concept-level labels, and propose selection strategies for student response histories. 
Our findings show that strategies based on conceptual alignment and semantic similarity significantly improve performance under limited context length. 
At this stage, we have only explored a limited set of selection strategies. 
Identifying more effective strategies is left for future work.
Additionally, predicting students' concept-level deficiencies remains challenging and still has significant room for improvement.

\section{Limitations}

This study adopts a single fixed prompt format and does not explore model-specific or diverse prompting strategies.
Relying on a single format may limit the model's potential to fully capture the complexity of students' learning processes.
The concept-level annotation in this study adopts a binary scheme, indicating whether a student demonstrated a deficiency in a particular concept, without capturing varying degrees of understanding.
However, in real-world educational scenarios, students often exhibit partial or context-dependent mastery rather than complete understanding or total deficiency.
Such coarse-grained labeling may oversimplify students' conceptual states and constrain the model's capacity to reason about nuanced patterns of conceptual understanding.

\section{Ethics Statement}

Our dataset did not contain any personal information, and all annotators were fully informed about the research purposes of the data prior to the annotation process.

Experts were informed that they could provide feedback if the predefined concept set was insufficient to capture the observed reasoning. No such cases were reported.

\section*{Acknowledgments}

This research was partially supported by National Science and Technology Council, Taiwan, under grant NSTC 114-2221-E-A49-057-MY3 and NSTC 114-2639-E-A49-001-ASP.

\section{Bibliographical References}\label{sec:reference}

\bibliographystyle{lrec2026-natbib}
\bibliography{lrec2026-example}

% \section{Language Resource References}
% \label{lr:ref}
\bibliographystylelanguageresource{lrec2026-natbib}
\bibliographylanguageresource{languageresource}

\end{document}